\documentclass[12pt]{article}
\usepackage{amsmath, amssymb}
\usepackage{algorithm}
\usepackage{algpseudocode}
\usepackage{graphicx}
\usepackage{natbib}    
\usepackage{url}       
\usepackage{xcolor}    
\usepackage{booktabs} 
\usepackage{multirow} 
\usepackage{caption}  
\usepackage{float}  
\usepackage[a4paper, top=1.5cm, bottom=1.5cm, left=2.5cm, right=2.5cm]{geometry}
\usepackage{hyperref}  

\hypersetup{
    colorlinks=true,      
    linkcolor=blue,      
    urlcolor=blue,        
    citecolor=blue        
}

\makeatletter
\def\@cite#1#2{\textcolor{blue}{[#1]}}   
\makeatother

\title{SALE-Based Offline Reinforcement Learning with Ensemble Q-Networks}
\author{zheng chun \\
        University of Science and Technology of China \\
        \texttt{zhengchun@mail.ustc.edu.cn}}
\date{}

\begin{document}

\maketitle

\begin{abstract}
\noindent 
In this work, we build upon the offline reinforcement learning algorithm TD7\citep{fujimoto2024sale},which incorporates State-Action Learned Embeddings (SALE) and a prioritized experience replay buffer (LAP), propose a model-free actor-critic algorithm that integrates ensemble Q-networks and a gradient diversity penalty from EDAC\citep{an2021uncertainty}. The ensemble Q-networks introduce penalties to guide the actor network toward in-distribution actions, effectively addressing the challenge of out-of-distribution actions. Meanwhile, the gradient diversity penalty encourages diverse Q-value gradients, further suppressing overestimation for out-of-distribution actions. Additionally, our method retains an adjustable behavior cloning (BC) term that directs the actor network toward dataset actions during early training stages, while gradually reducing its influence as the precision of the Q-ensemble improves. These enhancements work synergistically to improve the stability and precision of the training. Experimental results on the D4RL MuJoCo benchmarks demonstrate that our algorithm achieves higher convergence speed, stability, and performance compared to existing methods.
\end{abstract}

\section{Introduction}

Offline reinforcement learning (Offline RL) has gained popularity in recent years due to its ability to leverage pre-collected offline data without requiring interaction with the environment, thereby avoiding the need to simulate complex real-world environments. In reinforcement learning, agent policies are trained based on action value estimations. However, using online reinforcement learning methods only on offline data can lead to overestimation of the value of unknown actions (i.e. actions outside the dataset), resulting in suboptimal agent learning policies that may perform worse than those trained by supervised learning \citep{torabi2018behavioral, fu2020d4rl}. This issue typically necessitates the interaction with the real environment to be resolved. Another challenge faced by Offline RL is instability during training and low training efficiency. To address these problems, researchers have proposed methods such as conservative estimation of unknown actions, sequence modeling, and policy constraints \citep{kumar2020conservative, fujimoto2019benchmarking, chen2106decision, wang2022diffusion, prudencio2023survey, yang2022rorl}. However, many state-of-the-art algorithms suffer from excessively long training times and some require meticulous hyperparameter tuning.

To overcome these issues, TD3+BC \citep{fujimoto2021minimalist} proposed that by adaptively learning the behavior policy based on TD3 \citep{fujimoto2018addressing}, effective policies can be obtained. TD3 is well known for mitigating value overestimation through its twin Q networks and delayed policy updates. TD7 \citep{fujimoto2024sale} builds upon TD3 by adding mechanisms such as state-action learning encoders, prioritized replay buffers, and training checkpoints, achieving top performance on both online and offline reinforcement learning benchmarks. However, I have observed that TD7 still exhibits instability during training in offline datasets.

To address this problem while maintaining the simplicity inherent in TD3, we incorporated EDAC \citep{an2021uncertainty}, an ensemble Q-network approach. EDAC increases the diversity of Q-value gradients with respect to actions, reduces the number of Q-networks, and selects the minimum Q-value among the networks to penalize unknown actions, also achieving top performance in offline reinforcement learning tasks. However, EDAC still requires a large number of Q-networks, especially in the Hopper environment, and has excessively long training times, with the authors recommending $3 \times 10^6$ gradient updates.

Combining these two methods, we propose a new algorithm named EDTD7. In our experiments, we were pleasantly surprised to find that EDTD7 simultaneously resolves the training stability issues of TD7 and the problems of the large number of Q-networks and long training times in EDAC, while ultimately achieving excellent performance. Subsequently, through systematic ablation experiments, we demonstrated the impact of each component on performance, deepening our understanding of the various modifications.

\section{Related Work}

\textbf{Ensemble Networks.} Ensemble Q-networks are effective for penalizing unknown actions by simply increasing the number of Q-networks, leading to strong performance in offline reinforcement learning tasks. However, this approach suffers from low training efficiency and high computational resource requirements. To mitigate this issue, EDAC \citep{an2021uncertainty} introduces increased gradient diversity of Q-values with respect to actions, which helps address the problem effectively. Recently, research on ensemble networks has been progressing. For example, \citep{beeson2024balancing} systematically studied the impact of applying BC+ensemble to SAC \citep{haarnoja2018soft} and TD3, but the issue of training time remains unsolved. \citep{huang2024offline} proposed a method that constrains Q-networks by averaging the Q-values across the ensemble, which also showed good performance. Additionally, \citep{yang2022rorl} incorporated regularization of the policy and value function for states near the dataset while using ensemble Q-networks, achieving promising results but requiring substantial hyperparameter tuning. \citep{wangensemble} combined Q-ensemble networks with TD3 by outputting Q-values as the difference between the mean and the standard deviation of the ensemble Q-values, alongside a Bernoulli sampling-based BC term, yielding favorable experimental outcomes.

\textbf{Micro-Design Choices.} TD3+BC \citep{fujimoto2021minimalist} proposed a simple modification that has shown good results in offline reinforcement learning tasks. Many subsequent works have attempted similar designs to achieve comparable outcomes. \citep{nikulin2023anti} introduced Random Network Distillation (RND) and used Feature-wise Linear Modulation (FiLM) conditioning, achieving excellent performance. \citep{tarasov2024revisiting} integrated multiple design choices into a model-free algorithm and systematically studied the effects of different designs on performance. \citep{peng2023weighted} proposed a weighted policy constraint method to reduce the likelihood of learning suboptimal or inferior actions. However, this approach still faces stability issues during the training process.

\section{Background}

Reinforcement learning problems are typically framed within the framework of Markov Decision Processes (MDP)\citep{sutton2018reinforcement}. An MDP describes the scenario where an agent is in state \( s \) at a given time step, takes action \( a \), and interacts with the environment. Based on the transition model \( p(s' | s, a) \), the environment generates the next state \( s' \), and the agent receives a reward \( r \) based on the reward function \( R(r | s, a) \). The goal of the agent is to collect the cumulative discounted reward from its current state during the interaction with the environment. The cumulative discounted reward is defined as \( G_t = \sum_{k=0}^{\infty} \gamma^k r_{t+k+1} \), where \( G_t \) is the cumulative discounted reward starting from time step \( t \), \( \gamma \) is the discount factor (typically \( 0 \leq \gamma \leq 1 \)), representing the degree to which future rewards are discounted, and \( r_{t+k+1} \) is the reward obtained by the agent at time step \( t+k+1 \).In reinforcement learning, the MDP is described by a tuple \( (S, A, R, p, \gamma) \), where \( S \) is the state space, \( A \) is the action space, \( R \) is the reward function, \( p \) is the transition model, and \( \gamma \) is the discount factor. The goal of the agent is to learn a policy \( \pi \) that maximizes the cumulative discounted reward starting from the initial state \( s_0 \).

In reinforcement learning, both Q-networks and policy networks are commonly parameterized using neural networks to handle high-dimensional inputs and outputs. Q-networks approximate the action-value function \( Q(s, a) \) and are often implemented as ensemble networks to improve stability and uncertainty estimation. An ensemble of \( n \) Q-networks is parameterized as \( \{ Q_{\theta_1}, Q_{\theta_2}, \dots, Q_{\theta_n} \} \), where \( \theta_i \) represents the parameters of the \( i \)-th Q-network. The ensemble output is typically computed by taking the minimum value across all networks: \( Q_{\text{ensemble}}(s, a) = \min_{i=1}^{n} Q_{\theta_i}(s, a) \). Similarly, policy networks approximate the optimal action distribution given a state and are parameterized as \( \pi_{\varphi}(s) \), where \( \varphi \) denotes the parameters of the policy network.

\section{Method}
In this section, we will introduce what SALE is, explain how SALE integrates with the ensemble Q-network and the actor network, and finally present the overall loss function.

The objective of SALE is to learn a set of embeddings (\( z^{sa}, z^s \)) that capture relevant structures in the observation space as well as the transition dynamics of the environment. To achieve this, SALE employs a pair of parameterized encoders (\( f_{\theta_s}, g_{\theta_{sa}} \)): specifically, \( f_{\theta_s}(s) \) encodes the state \( s \) into the state embedding \( z^s \), while \( g_{\theta_{sa}}(z^s, a) \) jointly encodes the state embedding \( z^s \) and action \( a \) into the state-action embedding \( z^{sa} \):
\begin{equation}
z^s := f_{\theta_s}(s), \quad z^{sa} := g_{\theta_{sa}}(z^s, a) \label{eq:1}
\end{equation}
The embeddings are designed to capture the structural properties of the environment. However, they may not contain all the information required by the value and policy networks, such as features related to the reward function, the behavior policy, or the task horizon. To address this limitation, \citep{fujimoto2024sale} augment the embeddings by concatenating them with the original state and action. This allows the value and policy networks to learn richer and more relevant internal representations to better fulfill their respective tasks:
\begin{equation}
Q_{\theta_i}(s, a) \rightarrow Q_{\theta_i}(s, a, z^s, z^{sa}), \quad
\pi_{\varphi}(s) \rightarrow \pi_{\varphi}(s, z^s)
\end{equation}
To prevent instability in the embedding space, such as monotonic growth or collapse to a redundant representation, \citep{fujimoto2024sale} introduce a normalization layer called AvgL1Norm. This layer normalizes the input vector by dividing it by its average absolute value across all dimensions, ensuring the relative scale remains consistent throughout the learning process. Given an \( N \)-dimensional vector \( \mathbf{x} \), AvgL1Norm is defined as:
\begin{equation}
\text{AvgL1Norm}(\mathbf{x}) := \frac{\mathbf{x}}{\frac{1}{N} \sum_{i=1}^{N} |x_i|}
\end{equation}
where \( N \) is the dimensionality of \( \mathbf{x} \).

\begin{figure}[H]
    \centering
    \includegraphics[width=1.0\textwidth]{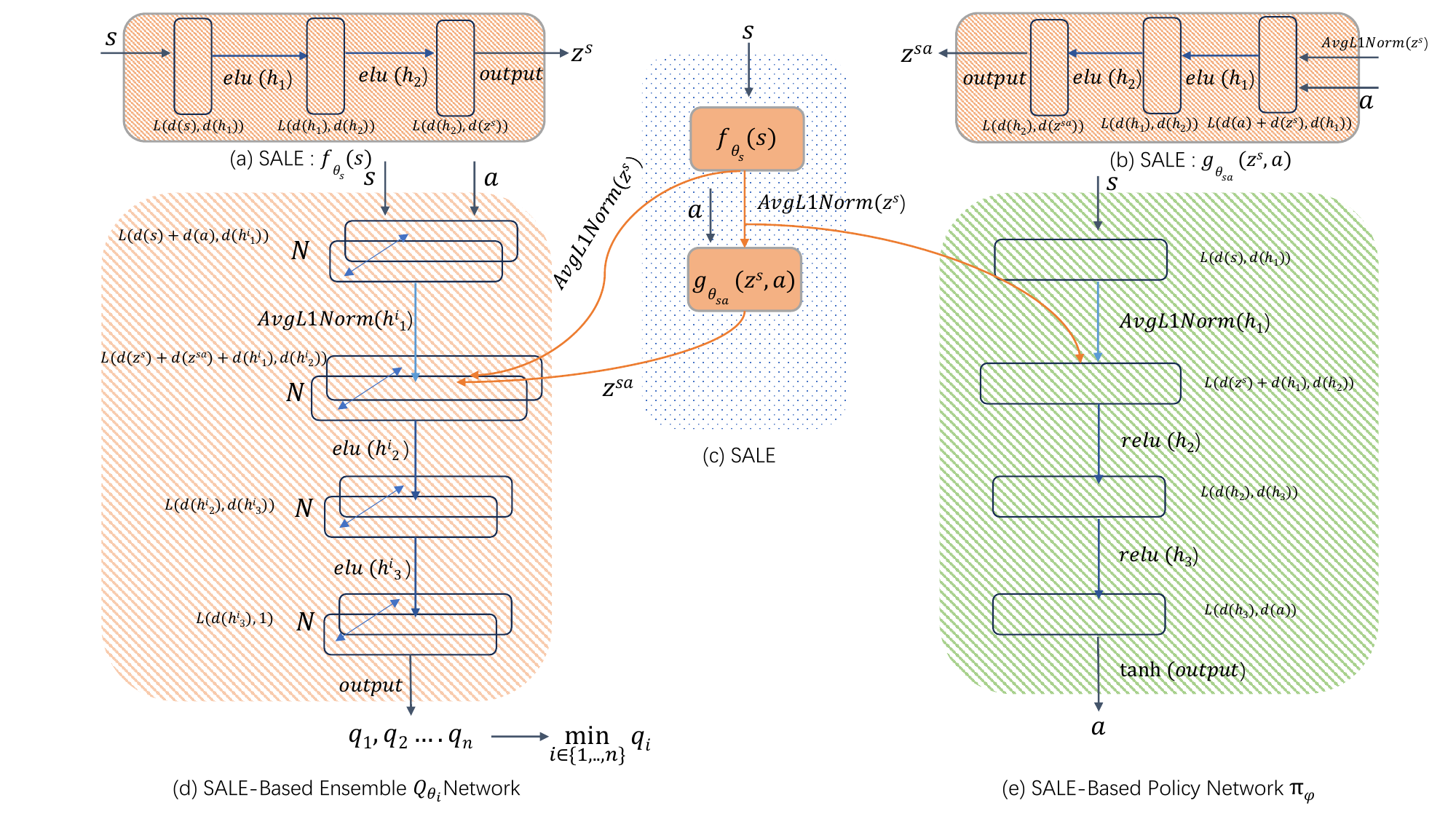}  
    \caption{\textbf{Overall network architecture.} The State-Action Learned Embeddings (SALE) method uses encoders \( \left( f_{\theta_s}, g_{\theta_{sa}} \right) \) to output the embeddings \( \left( z^s, z^{sa} \right) \) for states and state-action pairs, respectively. These embeddings are then passed through the corresponding hidden layers of the ensemble Q-network \( Q_{\theta_i} \) and the policy network \( \pi_{\varphi} \), where \( h_i \) denotes the output of the \( i \)-th hidden layer, and \( d(\cdot) \) represents the dimension of the input object (such as the state \( s \) or other objects). A linear transformation \( L(d(\cdot), d(\cdot)) \) is used to map the input dimensions to output dimensions in each hidden layer. After passing through the respective activation functions, the ensemble Q-network outputs the minimum value among the ensemble Q-values, while the policy network \( \pi_{\varphi} \) generates an action in the range of (-1, 1) after being processed by the \( \tanh \) activation function.}
    \label{fig:network_architecture}
\end{figure}

The overall network architecture and forward propagation path are shown in \hyperref[fig:network_architecture]{Figure~\ref{fig:network_architecture}}. During training, to address the overestimation of out-of-distribution actions, we select the minimum Q-value from the ensemble of Q-networks. Additionally, the method proposed by \citep{an2021uncertainty} is employed to reduce the gradient similarity between actions across different Q-networks. The resulting loss function for the ensemble Q-network is given by:
\begin{equation}
\begin{aligned}
\mathcal{L}_{Q} &= \mathbb{E}_{(s,a) \sim D} \left[ \text{huber} \left( \sum_{i=1}^{n} Q_{\theta_i}(s, a, z^s, z^{sa}) - \left( r + \gamma \hat{Q}_{\min} \right) \right) \right. \\
&\quad + \frac{\eta}{n-1} \sum_{1 \leq i \neq j \leq n} \text{ES}_{\theta_i, \theta_j}(s, a, z^s, z^{sa}) \Big]
\label{eq:QLoss}
\end{aligned}
\end{equation}
The Huber loss function is defined as:
\begin{equation}
\text{huber}(x) = \begin{cases} 
0.5 \cdot x^2 & \text{if } |x| < \text{min\_priority} \\
\text{min\_priority} \cdot |x| & \text{if } |x| \geq \text{min\_priority}
\end{cases}
\end{equation}
Additionally, the cosine similarity between the gradients of different Q-networks is used to enforce diversity in the actions taken by the networks:
\begin{equation}
\text{ES}_{\theta_i, \theta_j}(s, a, z^s, z^{sa}) = \cos\left( \nabla_a Q_{\theta_i}(s, a, z^s, z^{sa}), \nabla_a Q_{\theta_j}(s, a, z^s, z^{sa}) \right)
\end{equation}
The target network \( \hat{Q}_{\bar{\theta}_j} \) \citep{mnih2015human} helps prevent rapid changes in Q-values. To further stabilize training, the selected minimum Q-value from the ensemble is clipped between the minimum and maximum values of the target network's Q-values from the previous training iteration:
\begin{equation}
\hat{Q}_{\min} \approx \text{clip}\left( \min_{j \in \{1, 2, \dots, n\}} \hat{Q}_{\bar{\theta}_j}(s', \pi_{\overline{\varphi}}(s'), z^{s'}, z^{s'\pi_{\overline{\varphi}}(s')}), \min(\hat{Q}_{\min}^{\text{prev}}), \max(\hat{Q}_{\min}^{\text{prev}}) \right)
\label{eq:clip_formula}
\end{equation}
This clipping ensures that the selected minimum Q-value is within the range of the previous Q-values, thus preventing large deviations and maintaining stability in the Q-value updates.

The loss function for the policy network \( \pi_{\varphi} \) consists of two main components: the minimum Q-value from the ensemble of Q-networks and an adaptive behavior cloning (BC) term. The minimum Q-value helps mitigate the overestimation of out-of-distribution actions, while the BC term encourages the policy to replicate the target actions by minimizing the discrepancy between the predicted action and the target action. The resulting loss function is formulated as:
\begin{equation}
\begin{aligned}
\mathcal{L}_{\pi} &= \mathbb{E}_{(s, a) \sim D} \left[ - \min_{i \in \{1, 2, \dots, n\}} Q_{\theta_i}(s,\pi_{\varphi}(s), z^s, z^{s\pi_{\varphi}(s)}) \right] \\
&\quad + \lambda \cdot \mathbb{E}_{(s,a) \sim D} \left[ \left| \min_{i \in \{1, 2, \dots, n\}} Q_{\theta_i}(s,\pi_{\varphi}(s), z^s, z^{s\pi_{\varphi}(s)}) \right|_{\times} \cdot (\pi_{\varphi}(s, z^s) - a)^2 \right]
\end{aligned}
\label{eq:policy_loss}
\end{equation}
Here, the first part of the loss, \( - \min_{i \in \{1, 2, \dots, n\}} Q_{\theta_i}(s, a, z^s, z^{sa}) \), is the minimum Q-value from the ensemble of Q-networks, used to prevent overestimation of out-of-distribution actions. The second part, weighted by \( \lambda \), represents the adaptive behavior cloning (BC) term, which minimizes the discrepancy between the predicted action \( \pi_{\varphi}(s, z^s) \) and the target action \( a \). The term \( \left| (min_{i \in \{1, 2, \dots, n\}} Q_{\theta_i}(s,\pi_{\varphi}(s), z^s, z^{s\pi_{\varphi}(s)}) \right|_{\times} \) indicates that the absolute value of the Q-value is first taken, followed by gradient isolation (denoted by the cross symbol \( \times \)). This ensures that the gradient computation does not directly involve the Q-value, helping to stabilize the learning process by preventing the Q-value from affecting the gradient flow during backpropagation.

This loss function is designed to ensure that the policy network effectively learns from the offline data while maintaining stability, especially when addressing the challenge of out-of-distribution actions.

The encoder loss, denoted as \( \mathcal{L}_{\text{SALE}} \), is calculated as:
\begin{equation}
\mathcal{L}_{\text{SALE}} = \left( g_{\theta_{sa}}(f_{\theta_s}(s), a) - f_{\theta_s}(s')_{\times} \right)^2
\label{eq:encoder_loss}
\end{equation}
This loss is designed to learn the encoding for the next state by minimizing the squared difference between the predicted next state encoding and the current state encoding, with gradient isolation applied to the next state encoding, as indicated by the \( \times \) symbol.

In \hyperref[eq:clip_formula]{Equation~\ref{eq:clip_formula}}, \(a'\) is computed based on the TD3 method \citep{fujimoto2018addressing}, using the action generated by \( \pi_{\overline{\varphi}} \) (which is one version behind the most recently trained network) with added noise. Meanwhile, \(z^{s'}\) and \(z^{{s'}\pi_{\overline{\varphi}}(s')}\) are computed using \( f_{\overline{\overline{\theta}}_s} \) and \( g_{\overline{\overline{\theta}}_{sa}} \) (which are two versions behind the most recently trained network). In \hyperref[eq:QLoss]{Equation~\ref{eq:QLoss}}, other components compute \(z^s\) and \(z^{sa}\), relying only on \( f_{\overline{\theta}_s} \) and \( g_{\overline{\theta}_{sa}} \) (which are one version behind the most recently trained network). Similarly, in \hyperref[eq:policy_loss]{Equation~\ref{eq:policy_loss}}, the other components compute \(z^s\), \(z^{s\pi_{\varphi}(s)}\), relying solely on \( f_{\overline{\theta}_s} \) and \( g_{\overline{\theta}_{sa}} \). This strategy helps mitigate instability caused by inconsistent inputs \citep{fujimoto2024sale}.

Then,all target networks are updated using hard updates, i.e., every \( M \) gradient steps, the parameters of the target network are copied from the most recently trained network. The update rules are as follows:
\begin{equation}
\begin{aligned}
Q_{\overline{\theta}_i} &\leftarrow Q_{\theta_i}, \\
\pi_{\overline{\varphi}} &\leftarrow \pi_{\varphi}, \\
\left( f_{\overline{\overline{\theta}}_s}, g_{\overline{\overline{\theta}}_{sa}} \right) &\leftarrow \left( f_{\overline{\theta}_s}, g_{\overline{\theta}_{sa}} \right), \\
\left( f_{\overline{\theta}_s}, g_{\overline{\theta}_{sa}} \right) &\leftarrow \left( f_{\theta_s}, g_{\theta_{sa}} \right)
\end{aligned}
\label{eq:target_update}
\end{equation}
Finally, LAP \citep{horgan2018distributed,fujimoto2020equivalence} , following \citep{fujimoto2024sale}, the probability of sampling a sample from the replay buffer is given by the following formula:
\begin{equation}
p(i) = \frac{\max\left(|\delta(i)|^\alpha, 1\right)}{\sum_{j \in D} \max\left(|\delta(j)|^\alpha, 1\right)},
\quad \text{where } \delta(i) := Q(s, a) - y
\label{eq:sampling_probability}
\end{equation}
To summarize, the entire algorithm flow is shown in \hyperref[alg:EDTD7]{Algorithm~\ref{alg:EDTD7}}.

\begin{algorithm}[H]
\caption{EDTD7}
\label{alg:EDTD7}
\begin{algorithmic}[1]

\State \textbf{Initialize:} Load offline dataset \( D \)
\Statex \quad - Ensemble Q-networks \( \{ Q_{\theta_1}, Q_{\theta_2}, \dots, Q_{\theta_n} \} \),
\Statex \quad - Policy network \( \pi_{\varphi} \),
\Statex \quad - SALE network \( (f_{\theta_s}, g_{\theta_{sa}}) \),
\Statex \quad - Target networks \( \{ Q_{\overline{\theta}_1}, \dots, Q_{\overline{\theta}_n} \}, \pi_{\overline{\varphi}}, (f_{\overline{\theta}_s}, g_{\overline{\theta}_{sa}}), (f_{\overline{\overline{\theta}}_s}, g_{\overline{\overline{\theta}}_{sa}}) \),
\Statex \quad - Hyperparameters \( M \), \( E \), and \( \text{MAX} \)
\For{each training step \( t = 1 \) to \( MAX \)} 
    \State Sample batch \( (s, a, r, s') \sim D \) from LAP replay buffer \hyperref[eq:sampling_probability]{(Equation~\ref{eq:sampling_probability})}
    \State Train SALE minimizing \( \mathcal{L}_{\text{SALE}} \) \hyperref[eq:encoder_loss]{(Equation~\ref{eq:encoder_loss})}
    \State Train Q-networks minimizing \( \mathcal{L}_Q \) \hyperref[eq:QLoss]{(Equation~\ref{eq:QLoss})}
    \State Train policy network \( \pi_{\varphi} \) minimizing \( \mathcal{L}_{\pi} \) \hyperref[eq:policy_loss]{(Equation~\ref{eq:policy_loss})}
    \If{\( t \) mod \( M \) == 0}
        \State Update target networks \hyperref[eq:target_update]{(Equation~\ref{eq:target_update})}
    \EndIf
    \If{\( t \) mod \( E \) == 0}
        \State Interact with environment to compute evaluation score
    \EndIf
\EndFor

\end{algorithmic}
\end{algorithm}

\section{Experiment Results}
This section will focus on presenting the experimental results of EDTD7 in the D4RL MuJoCo benchmarks\citep{fu2020d4rl}, and provide a brief discussion and summary based on the results of the ablation studies.

\subsection{D4RL MuJoCo Benchmark}
The results of EDTD7, along with comparisons to other algorithms, are summarized in \hyperref[tab:result_base]{Table~\ref{tab:result_base}}. We compare EDTD7 with TD7\citep{fujimoto2024sale}, EDAC\citep{an2021uncertainty}, as well as several other representative and popular algorithms\citep{torabi2018behavioral,kumar2020conservative,kostrikov2021offline}. For each algorithm, we conduct local reproductions and run them for 1 million steps using 3 random seeds, evaluating the policy network during training every 5000 steps by interacting with the environment for 10 episodes. The complete learning curves can be found in \hyperref[learningCurves]{Appendix~\ref{learningCurves}}, which clearly demonstrates that our algorithm exhibits greater stability in comparison to others.In \hyperref[morebaseline]{Appendix~\ref{morebaseline}}, we further compare EDTD7 with other contemporary algorithms based on ensemble Q-networks, as well as some of the current state-of-the-art non-ensemble methods.

\subsection{EDTD7: Core Element Analysis}
\textbf{Hyperparameter Study.} We conducted experiments by varying the number of ensemble Q-networks \( N \), where \( N \in \{10, 20\} \), as well as the values of \( \eta \in \{0.0, 1.0, 5.0\} \) and the imitation coefficient \( \lambda \in \{0.0, 0.01, 0.05, 0.1\} \), with all results plotted in \hyperref[HyperparameterStudy]{Appendix~\ref{HyperparameterStudy}}. Notably, increasing the number of Q-networks does not necessarily improve performance, as excessively large penalties may not be suitable for certain datasets. For the \( \eta \) term, a value of 1.0 emerges as a stable choice, outperforming other settings, while for the \( \lambda \) term, although a value of 0.0 performs better in some datasets, \( \lambda = 0.01 \) generally achieves superior results in most cases. We hope this section provides valuable insights into the configuration of these hyperparameters.

\textbf{Other Target Value Choices.} We also experimented with an alternative target value, \textbf{pessQ}\citep{wangensemble}, for the ensemble Q-networks to compare with our current approach. For details, please refer to \hyperref[Target]{Appendix~\ref{Target}}. It can be observed that \textbf{pessQ} may also perform well in certain scenarios, but using our current \hyperref[eq:clip_formula]{Equation~\ref{eq:clip_formula}} demonstrates relatively stable and competitive performance.

\textbf{Component Ablation.} We conducted ablation studies on the three components: SALE, LAP, and the ensemble Q-networks. For details, please refer to \hyperref[Ablation]{Appendix~\ref{Ablation}}. Our results indicate that SALE consistently improves performance, while omitting LAP can be a viable option for certain datasets. However, removing the ensemble network significantly degrades performance. Overall, the combination of all three components achieves the best performance across most scenarios.

\begin{table}[H]
\centering
\caption{The reported scores represent the average D4RL normalized performance across the final 10 evaluations, computed over 3 random seeds, with ± standard deviation indicating seed-level variability. Our implementation of EDTD7, EDAC, and TD7 was evaluated on the D4RL MuJoCo environment “v4”\textsuperscript{\ref{fn:v4_env}}. For BC, CQL, and IQL, we relied on results from their original publications or re-implementations adhering to the authors’ suggested configurations. The best-performing method is emphasized, along with other approaches that achieve competitive results with lower variance.} 
\label{tab:result_base} 
\resizebox{\textwidth}{!}{ 
\begin{tabular}{lllllll}
\toprule
\textbf{Task} & \textbf{BC} & \textbf{CQL} & \textbf{EDAC} & \textbf{IQL} & \textbf{TD7} & \textbf{EDTD7(ours)} \\
\midrule
halfcheetah-random & 2.2$\pm$0.0 & 3.04$\pm$1.89 & 25.83$\pm$0.39 & 2.63$\pm$0.20 & 18.87$\pm$0.72 & \textbf{32.10$\pm$1.17} \\
halfcheetah-medium & 43.2$\pm$0.6 & 43.00$\pm$1.00 & 62.58$\pm$0.75 & 46.79$\pm$0.96 & 57.73$\pm$1.03 & \textbf{70.83$\pm$1.04} \\
halfcheetah-medium-replay & 37.6$\pm$2.1 & 19.01$\pm$20.66 & 55.25$\pm$2.42 & 41.56$\pm$8.17 & 54.01$\pm$1.03 & \textbf{61.62$\pm$2.71} \\
halfcheetah-medium-expert & 44.0$\pm$1.6 & 60.47$\pm$25.96 & 69.78$\pm$23.05 & 80.62$\pm$19.32 & \textbf{106.89$\pm$0.88} & 93.86$\pm$35.24 \\
halfcheetah-expert & 91.8$\pm$1.5 & 93.98$\pm$1.57 & 3.37$\pm$1.93 & 95.27$\pm$2.85 & 109.20$\pm$1.15 & \textbf{109.61$\pm$0.71} \\
halfcheetah-full-replay & 62.9$\pm$0.8 & 66.57$\pm$14.26 & 78.24$\pm$2.74 & 70.80$\pm$2.98 & 82.96$\pm$1.05 & \textbf{89.39$\pm$3.56} \\
\midrule
hopper-random & 3.7$\pm$0.6 & 8.16$\pm$1.15 & \textbf{31.45$\pm$0.04} & 7.26$\pm$0.52 & 12.01$\pm$4.40 & 17.07$\pm$10.31 \\
hopper-medium & 54.1$\pm$3.8 & 49.61$\pm$6.53 & \textbf{95.34$\pm$9.25} & 49.76$\pm$8.59 & 56.13$\pm$12.49 & 87.41$\pm$21.11 \\
hopper-medium-replay & 16.6$\pm$4.8 & 27.34$\pm$16.33 & 33.69$\pm$14.41 & 48.91$\pm$11.33 & 76.74$\pm$22.89 & \textbf{101.08$\pm$0.60} \\
hopper-medium-expert & 53.9$\pm$4.7 & 61.15$\pm$17.62 & 107.70$\pm$0.36 & 51.96$\pm$27.10 & \textbf{109.11$\pm$0.68} & 106.28$\pm$13.93 \\
hopper-expert & \textbf{107.7$\pm$9.7} & 96.35$\pm$19.33 & 96.75$\pm$21.80 & 57.16$\pm$19.24 & 73.33$\pm$24.67 & 107.63$\pm$1.38 \\
hopper-full-replay & 19.9$\pm$12.9 & 94.45$\pm$6.92 & 106.14$\pm$0.32 & 104.81$\pm$0.58 & 72.21$\pm$36.25 & \textbf{106.30$\pm$0.76} \\
\midrule
walker2d-random & 1.3$\pm$0.1 & 3.65$\pm$2.50 & 6.63$\pm$0.87 & 7.40$\pm$1.66 & 3.95$\pm$1.67 & \textbf{21.79$\pm$0.06} \\
walker2d-medium & 70.9$\pm$11.0 & 75.03$\pm$12.94 & 82.72$\pm$0.88 & 81.47$\pm$5.44 & 71.07$\pm$31.80 & \textbf{90.09$\pm$1.17} \\
walker2d-medium-replay & 20.3$\pm$9.8 & 49.40$\pm$27.36 & 81.77$\pm$0.73 & 60.87$\pm$31.23 & 68.42$\pm$37.74 & \textbf{85.87$\pm$2.99} \\
walker2d-medium-expert & 90.1$\pm$13.2 & 108.47$\pm$0.81 & 112.87$\pm$1.38 & 109.92$\pm$0.52 & 110.26$\pm$0.37 & \textbf{112.95$\pm$0.36} \\
walker2d-expert & 108.7$\pm$0.2 & 107.54$\pm$0.63 & 32.82$\pm$34.61 & 110.54$\pm$0.64 & 110.37$\pm$0.51 & \textbf{112.65$\pm$0.95} \\
walker2d-full-replay & 68.8$\pm$17.7 & 80.52$\pm$16.31 & 94.22$\pm$0.88 & 93.02$\pm$1.52 & 97.98$\pm$0.98 & \textbf{98.05$\pm$1.85} \\
\bottomrule
\end{tabular}
}
\end{table}
\footnotetext[1]{\label{fn:v4_env}We chose the v4 environment because it is the latest version, easier to deploy, and more widely adopted for benchmarking. The v2 environment is older, and all algorithms were tested in the same environment to ensure fairness and avoid bias.}
\section{Conclusion}
In this work, building upon TD7 and EDAC, we introduce an enhanced offline reinforcement learning algorithm, EDTD7, which integrates ensemble Q-networks and a novel combination of components to improve performance on the D4RL MuJoCo benchmarks. By leveraging ensemble Q-networks, we construct a robust and stable value estimation framework, which significantly reduces variance during evaluation and enhances overall performance. Additionally, we conduct extensive ablation studies to analyze the contributions of key components, including SALE, LAP, and the ensemble Q-networks, demonstrating that their combination leads to substantial performance improvements.

One of the key strengths of EDTD7 is its simplicity and efficiency. Despite its straightforward implementation, EDTD7 achieves competitive results compared to state-of-the-art offline RL algorithms, while maintaining low computational overhead. The algorithm’s ability to balance exploration and exploitation through ensemble-based uncertainty estimation is a critical factor in its success. Furthermore, our hyperparameter study provides insights into the optimal configuration of ensemble size, regularization terms, and imitation coefficients, offering practical guidance for future applications.

A notable limitation of EDTD7 is the need to preset certain hyperparameters, such as the ensemble size \( N \) and the imitation coefficient \( \lambda \). While these parameters can be tuned through empirical studies, their optimal values may vary across different tasks and datasets. However, this challenge is not unique to our approach and is a common issue in offline RL. To address this, we provide a comprehensive analysis of hyperparameter sensitivity, which can serve as a valuable reference for practitioners.

Looking ahead, one potential extension of this work could involve automating the selection of hyperparameters, such as through meta-learning or multi-armed bandit approaches. Additionally, further exploration of the interplay between ensemble methods and behavior cloning could yield new insights into improving offline RL algorithms. By leveraging the inherent structure and characteristics of offline datasets, we believe that EDTD7 can inspire the development of simpler, yet highly effective, algorithms that prioritize stability, interpretability, and performance.

In conclusion, EDTD7 represents a significant step forward in offline reinforcement learning, demonstrating that a well-designed combination of ensemble methods and multiple components (including SALE, LAP, etc. in addition to BC) can achieve state-of-the-art results while maintaining simplicity and efficiency. We hope that this work encourages further research into leveraging ensemble techniques and offline dataset characteristics to develop robust and practical offline RL algorithms.

\bibliographystyle{plainnat}  
\bibliography{main}   

\begin{thebibliography}{28}
\providecommand{\natexlab}[1]{#1}
\providecommand{\url}[1]{\texttt{#1}}
\expandafter\ifx\csname urlstyle\endcsname\relax
  \providecommand{\doi}[1]{doi: #1}\else
  \providecommand{\doi}{doi: \begingroup \urlstyle{rm}\Url}\fi

\bibitem[An et~al.(2021)An, Moon, Kim, and Song]{an2021uncertainty}
Gaon An, Seungyong Moon, Jang-Hyun Kim, and Hyun~Oh Song.
\newblock Uncertainty-based offline reinforcement learning with diversified q-ensemble.
\newblock \emph{Advances in neural information processing systems}, 34:\penalty0 7436--7447, 2021.

\bibitem[Beeson and Montana(2024)]{beeson2024balancing}
Alex Beeson and Giovanni Montana.
\newblock Balancing policy constraint and ensemble size in uncertainty-based offline reinforcement learning.
\newblock \emph{Machine Learning}, 113\penalty0 (1):\penalty0 443--488, 2024.

\bibitem[Brockman(2016)]{brockman2016openai}
G~Brockman.
\newblock Openai gym.
\newblock \emph{arXiv preprint arXiv:1606.01540}, 2016.

\bibitem[Chen et~al.(2021)Chen, Lu, Rajeswaran, Lee, Grover, Laskin, Abbeel, Srinivas, and Mordatch]{chen2106decision}
Lili Chen, Kevin Lu, Aravind Rajeswaran, Kimin Lee, Aditya Grover, Michael Laskin, Pieter Abbeel, Aravind Srinivas, and Igor Mordatch.
\newblock Decision transformer: reinforcement learning via sequence modeling, 2 june 2021.
\newblock \emph{URL http://arxiv.org/abs/2106.01345}, 2021.

\bibitem[Fu et~al.(2020)Fu, Kumar, Nachum, Tucker, and Levine]{fu2020d4rl}
Justin Fu, Aviral Kumar, Ofir Nachum, George Tucker, and Sergey Levine.
\newblock D4rl: Datasets for deep data-driven reinforcement learning.
\newblock \emph{arXiv preprint arXiv:2004.07219}, 2020.

\bibitem[Fujimoto and Gu(2021)]{fujimoto2021minimalist}
Scott Fujimoto and Shixiang~Shane Gu.
\newblock A minimalist approach to offline reinforcement learning.
\newblock \emph{Advances in neural information processing systems}, 34:\penalty0 20132--20145, 2021.

\bibitem[Fujimoto et~al.(2018)Fujimoto, Hoof, and Meger]{fujimoto2018addressing}
Scott Fujimoto, Herke Hoof, and David Meger.
\newblock Addressing function approximation error in actor-critic methods.
\newblock In \emph{International conference on machine learning}, pages 1587--1596. PMLR, 2018.

\bibitem[Fujimoto et~al.(2019)Fujimoto, Conti, Ghavamzadeh, and Pineau]{fujimoto2019benchmarking}
Scott Fujimoto, Edoardo Conti, Mohammad Ghavamzadeh, and Joelle Pineau.
\newblock Benchmarking batch deep reinforcement learning algorithms.
\newblock \emph{arXiv preprint arXiv:1910.01708}, 2019.

\bibitem[Fujimoto et~al.(2020)Fujimoto, Meger, and Precup]{fujimoto2020equivalence}
Scott Fujimoto, David Meger, and Doina Precup.
\newblock An equivalence between loss functions and non-uniform sampling in experience replay.
\newblock \emph{Advances in neural information processing systems}, 33:\penalty0 14219--14230, 2020.

\bibitem[Fujimoto et~al.(2024)Fujimoto, Chang, Smith, Gu, Precup, and Meger]{fujimoto2024sale}
Scott Fujimoto, Wei-Di Chang, Edward Smith, Shixiang~Shane Gu, Doina Precup, and David Meger.
\newblock For sale: State-action representation learning for deep reinforcement learning.
\newblock \emph{Advances in Neural Information Processing Systems}, 36, 2024.

\bibitem[Haarnoja et~al.(2018)Haarnoja, Zhou, Abbeel, and Levine]{haarnoja2018soft}
Tuomas Haarnoja, Aurick Zhou, Pieter Abbeel, and Sergey Levine.
\newblock Soft actor-critic: Off-policy maximum entropy deep reinforcement learning with a stochastic actor.
\newblock In \emph{International conference on machine learning}, pages 1861--1870. PMLR, 2018.

\bibitem[Horgan et~al.(2018)Horgan, Quan, Budden, Barth-Maron, Hessel, Van~Hasselt, and Silver]{horgan2018distributed}
Dan Horgan, John Quan, David Budden, Gabriel Barth-Maron, Matteo Hessel, Hado Van~Hasselt, and David Silver.
\newblock Distributed prioritized experience replay.
\newblock \emph{arXiv preprint arXiv:1803.00933}, 2018.

\bibitem[Huang et~al.(2024)Huang, Dong, Pang, Liu, and Zhang]{huang2024offline}
Longyang Huang, Botao Dong, Ning Pang, Ruonan Liu, and Weidong Zhang.
\newblock Offline reinforcement learning without regularization and pessimism.
\newblock 2024.

\bibitem[Kingma(2014)]{kingma2014adam}
Diederik~P Kingma.
\newblock Adam: A method for stochastic optimization.
\newblock \emph{arXiv preprint arXiv:1412.6980}, 2014.

\bibitem[Kostrikov et~al.(2021)Kostrikov, Nair, and Levine]{kostrikov2021offline}
Ilya Kostrikov, Ashvin Nair, and Sergey Levine.
\newblock Offline reinforcement learning with implicit q-learning.
\newblock \emph{arXiv preprint arXiv:2110.06169}, 2021.

\bibitem[Kumar et~al.(2020)Kumar, Zhou, Tucker, and Levine]{kumar2020conservative}
Aviral Kumar, Aurick Zhou, George Tucker, and Sergey Levine.
\newblock Conservative q-learning for offline reinforcement learning.
\newblock \emph{Advances in Neural Information Processing Systems}, 33:\penalty0 1179--1191, 2020.

\bibitem[Mnih et~al.(2015)Mnih, Kavukcuoglu, Silver, Rusu, Veness, Bellemare, Graves, Riedmiller, Fidjeland, Ostrovski, et~al.]{mnih2015human}
Volodymyr Mnih, Koray Kavukcuoglu, David Silver, Andrei~A Rusu, Joel Veness, Marc~G Bellemare, Alex Graves, Martin Riedmiller, Andreas~K Fidjeland, Georg Ostrovski, et~al.
\newblock Human-level control through deep reinforcement learning.
\newblock \emph{nature}, 518\penalty0 (7540):\penalty0 529--533, 2015.

\bibitem[Nikulin et~al.(2023)Nikulin, Kurenkov, Tarasov, and Kolesnikov]{nikulin2023anti}
Alexander Nikulin, Vladislav Kurenkov, Denis Tarasov, and Sergey Kolesnikov.
\newblock Anti-exploration by random network distillation.
\newblock In \emph{International Conference on Machine Learning}, pages 26228--26244. PMLR, 2023.

\bibitem[Paszke et~al.(2019)Paszke, Gross, Massa, Lerer, Bradbury, Chanan, Killeen, Lin, Gimelshein, Antiga, et~al.]{paszke2019pytorch}
Adam Paszke, Sam Gross, Francisco Massa, Adam Lerer, James Bradbury, Gregory Chanan, Trevor Killeen, Zeming Lin, Natalia Gimelshein, Luca Antiga, et~al.
\newblock Pytorch: An imperative style, high-performance deep learning library.
\newblock \emph{Advances in neural information processing systems}, 32, 2019.

\bibitem[Peng et~al.(2023)Peng, Han, Liu, and Zhou]{peng2023weighted}
Zhiyong Peng, Changlin Han, Yadong Liu, and Zongtan Zhou.
\newblock Weighted policy constraints for offline reinforcement learning.
\newblock In \emph{Proceedings of the AAAI Conference on Artificial Intelligence}, volume~37, pages 9435--9443, 2023.

\bibitem[Prudencio et~al.(2023)Prudencio, Maximo, and Colombini]{prudencio2023survey}
Rafael~Figueiredo Prudencio, Marcos~ROA Maximo, and Esther~Luna Colombini.
\newblock A survey on offline reinforcement learning: Taxonomy, review, and open problems.
\newblock \emph{IEEE Transactions on Neural Networks and Learning Systems}, 2023.

\bibitem[Sutton(2018)]{sutton2018reinforcement}
Richard~S Sutton.
\newblock Reinforcement learning: An introduction.
\newblock \emph{A Bradford Book}, 2018.

\bibitem[Tarasov et~al.(2024)Tarasov, Kurenkov, Nikulin, and Kolesnikov]{tarasov2024revisiting}
Denis Tarasov, Vladislav Kurenkov, Alexander Nikulin, and Sergey Kolesnikov.
\newblock Revisiting the minimalist approach to offline reinforcement learning.
\newblock \emph{Advances in Neural Information Processing Systems}, 36, 2024.

\bibitem[Todorov et~al.(2012)Todorov, Erez, and Tassa]{todorov2012mujoco}
Emanuel Todorov, Tom Erez, and Yuval Tassa.
\newblock Mujoco: A physics engine for model-based control.
\newblock In \emph{2012 IEEE/RSJ international conference on intelligent robots and systems}, pages 5026--5033. IEEE, 2012.

\bibitem[Torabi et~al.(2018)Torabi, Warnell, and Stone]{torabi2018behavioral}
Faraz Torabi, Garrett Warnell, and Peter Stone.
\newblock Behavioral cloning from observation.
\newblock \emph{arXiv preprint arXiv:1805.01954}, 2018.

\bibitem[Wang and Zhang(2024)]{wangensemble}
Danyang Wang and Lingsong Zhang.
\newblock Ensemble-based offline reinforcement learning with adaptive behavior cloning.
\newblock In \emph{Adaptive Foundation Models: Evolving AI for Personalized and Efficient Learning}, 2024.

\bibitem[Wang et~al.(2022)Wang, Hunt, and Zhou]{wang2022diffusion}
Zhendong Wang, Jonathan~J Hunt, and Mingyuan Zhou.
\newblock Diffusion policies as an expressive policy class for offline reinforcement learning.
\newblock \emph{arXiv preprint arXiv:2208.06193}, 2022.

\bibitem[Yang et~al.(2022)Yang, Bai, Ma, Wang, Zhang, and Han]{yang2022rorl}
Rui Yang, Chenjia Bai, Xiaoteng Ma, Zhaoran Wang, Chongjie Zhang, and Lei Han.
\newblock Rorl: Robust offline reinforcement learning via conservative smoothing.
\newblock \emph{Advances in neural information processing systems}, 35:\penalty0 23851--23866, 2022.

\end{thebibliography}

\newpage  

\appendix  
\section{EDTD7 Additional Details}

\subsection{Experimental Details}

We use the following software versions:
\begin{itemize}
    \item \textbf{Python 3.9.18}
    \item \textbf{PyTorch 2.2.1} \citep{paszke2019pytorch}
    \item \textbf{CUDA Version 12.1}
    \item \textbf{Gym 0.26.2} \citep{brockman2016openai}
    \item \textbf{MuJoCo 3.2.5} \citep{todorov2012mujoco}
\end{itemize}

For all experiments, we use the ``v2'' version of the D4RL datasets, where a normalized D4RL score is provided. Formally, the D4RL score is defined as:
\[
\text{D4RL Score} = 100 \times \frac{\text{score} - \text{random score}}{\text{expert score} - \text{random score}}.
\]
We run on the following hardware:
\begin{itemize}
    \item \textbf{GPU: RTX 2080 Ti (11GB) $\times$ 1}
    \item \textbf{CPU: 12 vCPUs Intel(R) Xeon(R) Platinum 8255C CPU @ 2.50GHz}
\end{itemize}
The average wall clock runtime for the EDTD7 algorithm with 1 million updates is approximately \textbf{3.3 hours}, while TD7 takes around \textbf{2.5 hours}.

We use the following generic hyperparameters in our experiments, as shown in \hyperref[tab:hyperparameters]{Table~\ref{tab:hyperparameters}},overall network architecture as shown in \hyperref[fig:network_architecture]{Figure~\ref{fig:network_architecture}}.
As shown in \hyperref[tab:environment_hyperparameters]{Table~\ref{tab:environment_hyperparameters}}, we specify the ensemble size N, \(\eta\), and BC weight (\(\lambda\)) for each environment.
\begin{table}[H]
\centering
\begin{tabular}{@{} l l l @{}}
\toprule
\textbf{Component} & \textbf{Hyperparameter} & \textbf{Value} \\
\midrule
\multirow{5}{*}{\textbf{Common}} 
 & Discount factor $\gamma$ & 0.99 \\
 & Mini-batch size & 256 \\
 & Target update frequency & 250 \\
 & Optimizer (Shared) & Adam \citep{kingma2014adam} \\
 & Learning rate (Shared) & $3 \times 10^{-4}$ \\
\addlinespace
\multirow{3}{*}{\textbf{TD3 \citep{fujimoto2018addressing}}} 
 & Target policy noise $\sigma$ & $N(0, 0.2)$ \\
 & Target policy noise clipping $c$ & $(-0.5, 0.5)$ \\
 & Policy update frequency & 2 \\
\addlinespace
\multirow{2}{*}{\textbf{LAP \citep{fujimoto2020equivalence}}} 
 & Probability smoothing $\alpha$ & 0.4 \\
 & Minimum priority & 1 \\
\addlinespace
\multirow{2}{*}{\textbf{Encoder Model}} 
 & Encoder hidden dimension & 256 \\
 & Encoder activation & ELU \\
\addlinespace
\multirow{2}{*}{\textbf{Actor Model}} 
 & Actor hidden dimension & 256 \\
 & Actor activation & ReLU \\
\addlinespace
\multirow{2}{*}{\textbf{Q Ensemble}} 
 & Q activation & ELU \\
 & Q hidden dimension & 256 \\
\bottomrule
\end{tabular}
\caption{Hyperparameters used in the experiments.}
\label{tab:hyperparameters}
\end{table}

\begin{table}[H]
\centering
\begin{tabular}{@{} l l c c c @{}}
\toprule
\textbf{Environment} & \textbf{Task} & \textbf{N} & \(\eta\) & \(\lambda\) \\
\midrule
\multirow{6}{*}{\textbf{HalfCheetah}} 
 & halfcheetah-random        & 10 & 1 & 0.01 \\
 & halfcheetah-medium        & 10 & 1 & 0.01 \\
 & halfcheetah-medium-replay & 10 & 1 & 0.01 \\
 & halfcheetah-medium-expert & 10 & 1 & 0.01 \\
 & halfcheetah-expert        & 10 & 1 & 0.01 \\
 & halfcheetah-full-replay   & 10 & 1 & 0.01 \\
\addlinespace
\multirow{6}{*}{\textbf{Hopper}} 
 & hopper-random             & 10 & 1 & 0.01 \\
 & hopper-medium             & 10 & 1 & 0.01 \\
 & hopper-medium-replay      & 10 & 1 & 0.01 \\
 & hopper-medium-expert      & 10 & 1 & 0.05 \\
 & hopper-expert             & 10 & 1 & 0.05 \\
 & hopper-full-replay        & 10 & 1 & 0.01 \\
\addlinespace
\multirow{6}{*}{\textbf{Walker2d}} 
 & walker2d-random           & 10 & 1 & 0.01 \\
 & walker2d-medium           & 10 & 1 & 0.01 \\
 & walker2d-medium-replay    & 10 & 1 & 0.01 \\
 & walker2d-medium-expert    & 10 & 1 & 0.01 \\
 & walker2d-expert           & 10 & 1 & 0.01 \\
 & walker2d-full-replay      & 10 & 1 & 0.01 \\
\bottomrule
\end{tabular}
\caption{Per-environment hyperparameters: ensemble size (\(N\)), \(\eta\), and BC weight (\(\lambda\)).}
\label{tab:environment_hyperparameters}
\end{table}

\subsection{Learning Curves}
\label{learningCurves}

\begin{figure}[H]
    \centering
    \includegraphics[width=1.0\textwidth]{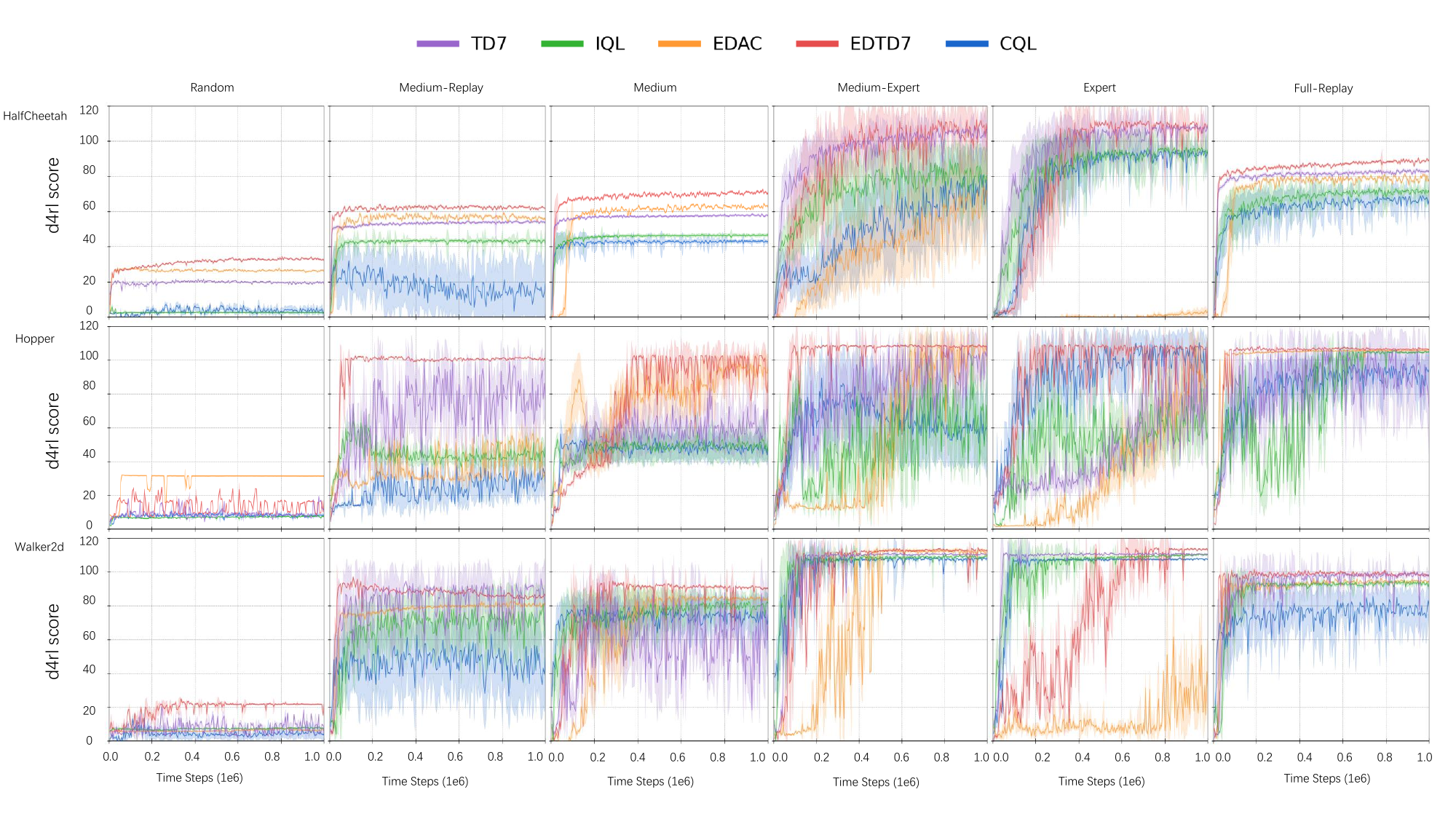}  
    \caption{The learning curves illustrate the performance of EDTD7 in comparison to TD7, EDAC, IQL, and CQL. Each curve represents the average performance over 4 random seeds, with the shaded regions indicating the standard deviation across these seeds.}
    \label{fig:learning_curves_comparison}
\end{figure}

\subsection{More Baseline Comparison}
\label{morebaseline}

We further expand our comparative analysis by incorporating the latest state-of-the-art baseline methods on the D4RL MuJoCo datasets. This includes ensemble-based approaches such as EDAC, RORL, and EABC, as well as innovative methods like SAC-RND and ReBRAC, which feature unique and sophisticated designs\citep{an2021uncertainty,yang2022rorl,wangensemble,nikulin2023anti,tarasov2024revisiting}.The results are either derived from the original publications or replicated using the official implementations provided by the authors. The comprehensive findings are summarized in \hyperref[tab:other_baseline]{Table~\ref{tab:other_baseline}}.

\begin{table}[H]
\centering
\caption{Average D4RL normalized score over the final 10 evaluations on 3 seeds, ± standard deviation over seeds. Experiment results for other algorithms are based on their original paper, or re-implements with author’s recommended parameters.We highlight the best average.} 
\label{tab:other_baseline} 
\resizebox{\textwidth}{!}{ 
\begin{tabular}{lllllll}
\toprule
\textbf{Task} & \textbf{EDAC} & \textbf{SAC-RND} & \textbf{ReBRAC} & \textbf{RORL} & \textbf{EABC} & \textbf{EDTD7} \\
              & \textbf{(papers)} & & & & & \textbf{(ours)} \\
\midrule
halfcheetah-random & 28.4$\pm$1.0 & 27.6$\pm$2.1 & 29.5$\pm$1.5 & 28.5$\pm$0.8 & \textbf{32.4$\pm$0.7} & 32.10$\pm$1.17 \\
halfcheetah-medium & 65.9$\pm$0.6 & 66.4$\pm$1.4 & 65.9$\pm$1.0 & 66.8$\pm$0.7 & 67.3$\pm$0.90 & \textbf{70.83$\pm$1.04} \\
halfcheetah-medium-replay & 61.3$\pm$1.9 & 51.2$\pm$3.2 & 51.0$\pm$0.8 & 61.9$\pm$1.5 & 61.4$\pm$1.6 & \textbf{61.62$\pm$2.71} \\
halfcheetah-medium-expert & 106.3$\pm$1.9 & \textbf{108.1$\pm$1.5} & 101.1$\pm$5.2 & 107.8$\pm$1.1 & 92.9$\pm$1.9 & 93.86$\pm$35.24 \\
halfcheetah-expert & 106.8$\pm$3.4 & 102.6$\pm$4.2 & 105.9$\pm$1.7 & 105.2$\pm$0.7 & 97.6$\pm$0.2 & \textbf{109.61$\pm$0.71} \\
\midrule
hopper-random & 25.3$\pm$10.4 & 19.6$\pm$12.4 & 8.1$\pm$2.4 & \textbf{31.4$\pm$0.1} & 31.5$\pm$0.4 & 17.07$\pm$10.31 \\
hopper-medium & 101.0$\pm$0.5 & 91.1$\pm$10.1 & 102.0$\pm$1.0 & \textbf{104.8$\pm$0.1} & 92.4$\pm$3.9 & 87.41$\pm$21.11 \\
hopper-medium-replay & 101.0$\pm$0.5 & 97.2$\pm$9.0 & 98.1$\pm$5.3 & \textbf{102.8$\pm$0.5} & 102.6$\pm$1.4 & 101.08$\pm$0.60 \\
hopper-medium-expert & 110.7$\pm$0.1 & 109.8$\pm$0.6 & 107.0$\pm$6.4 & \textbf{112.7$\pm$0.2} & 104$\pm$3.6 & 106.28$\pm$13.93 \\
hopper-expert & 110.1$\pm$0.1 & 109.8$\pm$0.5 & 100.1$\pm$8.3 & \textbf{112.8$\pm$0.2} & 111.2$\pm$0.3 & 107.63$\pm$1.38 \\
\midrule
walker2d-random & 16.6$\pm$7.0 & 18.7$\pm$6.9 & 18.4$\pm$4.5 & \textbf{21.4$\pm$0.2} & 1.7$\pm$1.7 & 21.79$\pm$0.06 \\
walker2d-medium & 92.5$\pm$0.8 & 92.7$\pm$1.2 & 82.5$\pm$3.6 & \textbf{102.4$\pm$1.4} & 89.0$\pm$0.6 & 90.09$\pm$1.17 \\
walker2d-medium-replay & 87.1$\pm$2.3 & 89.4$\pm$3.8 & 77.3$\pm$7.9 & 90.4$\pm$0.5 & \textbf{93.2$\pm$2.9} & 85.87$\pm$2.99 \\
walker2d-medium-expert & 114.7$\pm$0.9 & 104.6$\pm$11.2 & 111.6$\pm$0.3 & \textbf{121.2$\pm$1.5} & 112.0$\pm$0.3 & 112.95$\pm$0.36 \\
walker2d-expert & 115.1$\pm$1.9 & 104.5$\pm$22.8 & 112.3$\pm$0.2 & \textbf{115.4$\pm$0.5} & 110.8$\pm$0.1 & 112.65$\pm$0.95 \\
\bottomrule
\end{tabular} 
}
\end{table}

\section{EDTD7: Core Element Analysis}

\subsection{Hyperparameter Study}
\label{HyperparameterStudy}

This section presents a study of hyperparameters, where each experiment follows a controlled variable approach. Specifically, only the target hyperparameter is varied, while all other hyperparameters remain consistent with the settings provided in \hyperref[tab:environment_hyperparameters]{Table~\ref{tab:environment_hyperparameters}}.

\begin{figure}[H]
    \centering
    \includegraphics[width=1.0\textwidth]{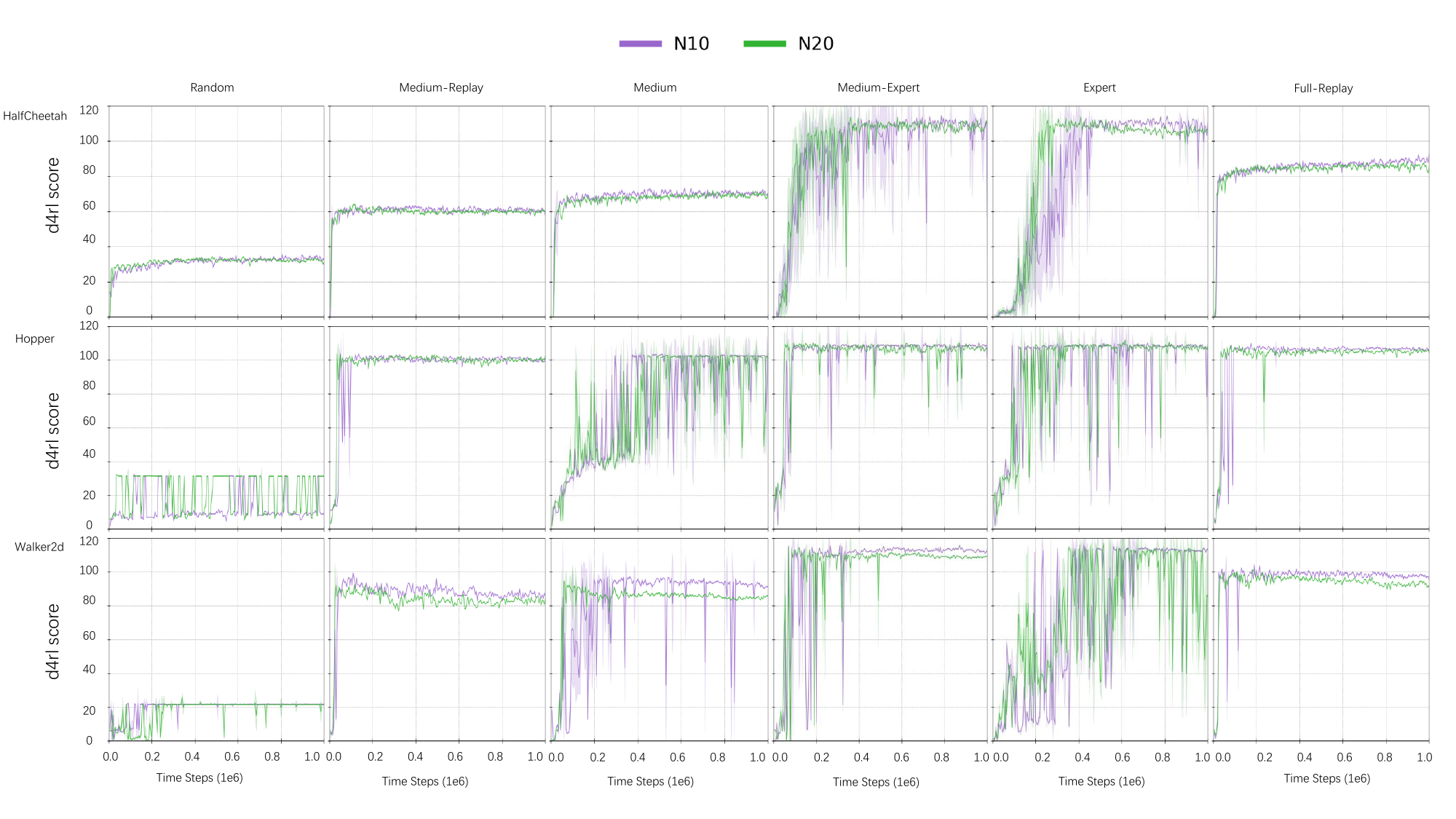}  
    \caption{The learning curves illustrate the performance of EDTD7 with varying ensemble sizes. Each curve represents the average performance, and the shaded regions indicate the standard deviation.}
    \label{fig:learning_curves_comparison_NQ}
\end{figure}

\begin{figure}[H]
    \centering
    \includegraphics[width=1.0\textwidth]{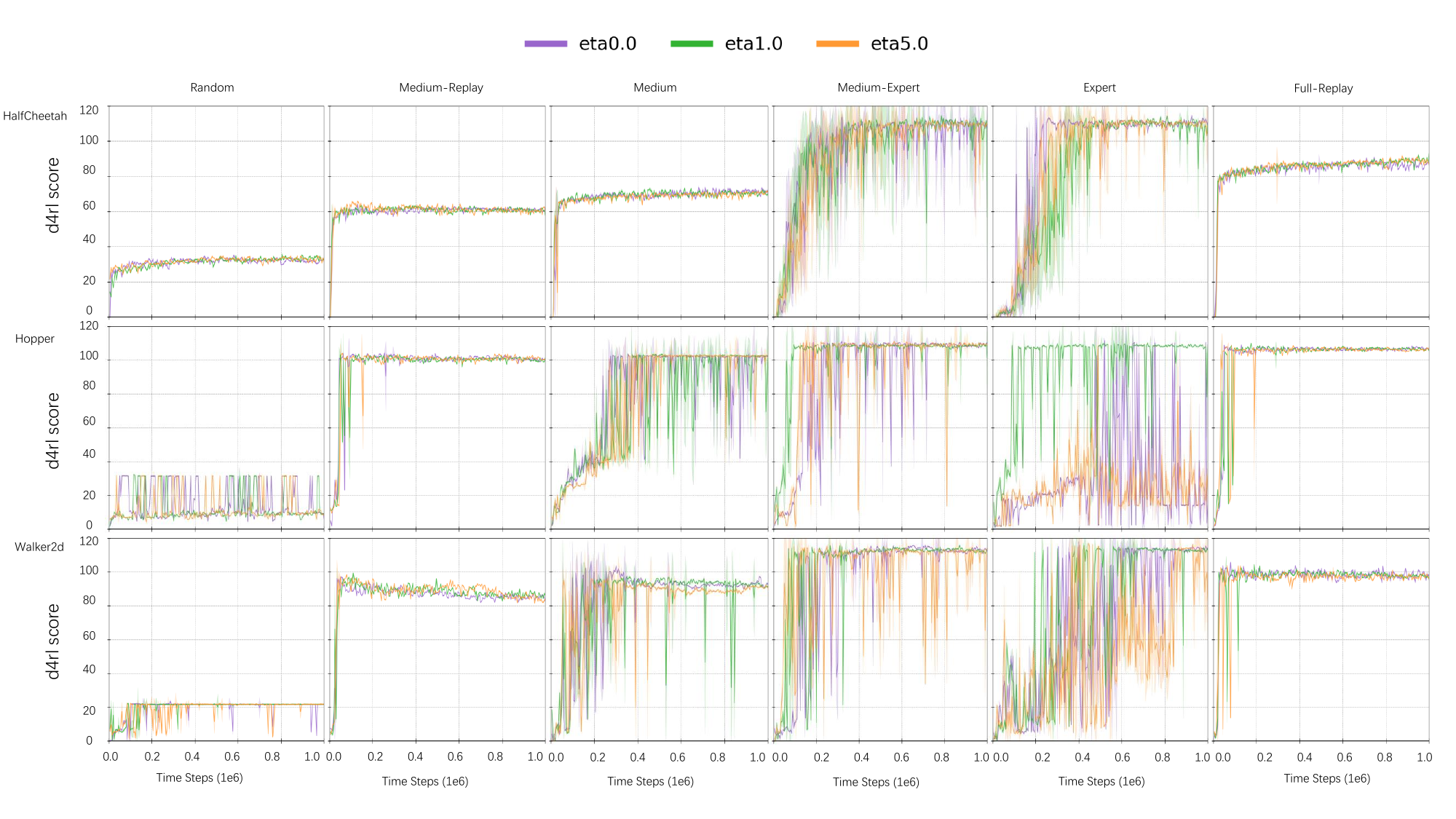}  
    \caption{The learning curves illustrate the impact of the $\eta$ term (Q-value gradient diversity penalty) in EDTD7. Each curve represents the average performance for different values of $\eta$, where a larger $\eta$ indicates a stronger penalty and higher gradient diversity. The shaded regions indicate the standard deviation.}
    \label{fig:learning_curves_comparison_eta}
\end{figure}

\begin{figure}[H]
    \centering
    \includegraphics[width=1.0\textwidth]{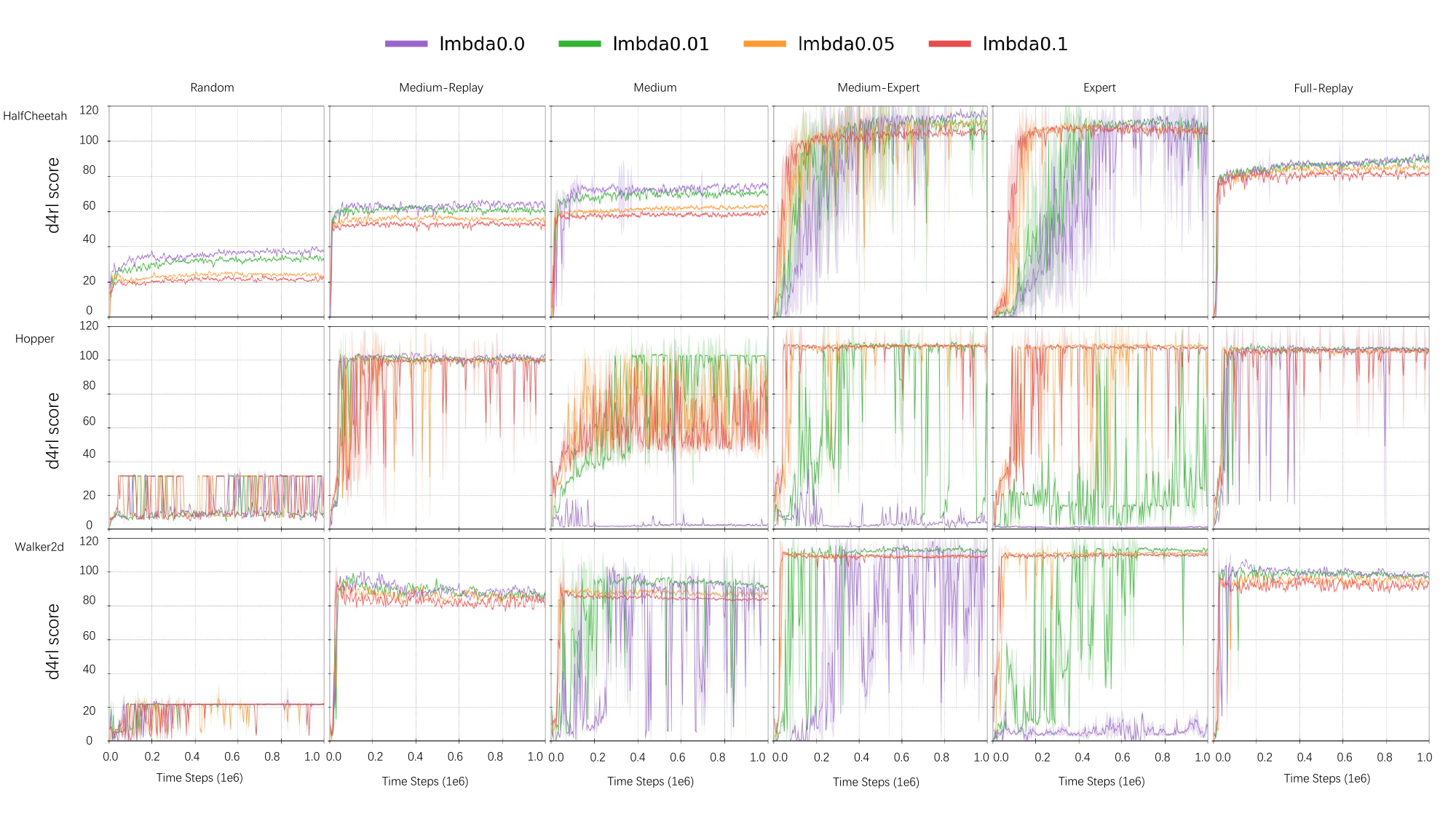}  
    \caption{The learning curves illustrate the impact of the $\lambda$ term (imitation learning strength) in EDTD7. Each curve represents the average performance for different values of $\lambda$, where a larger $\lambda$ indicates a stronger emphasis on imitation learning. The shaded regions indicate the standard deviation.}
    \label{fig:learning_curves_comparison_lmbda}
\end{figure}

\subsection{Other Target Value Choices}
\label{Target}

This subsection investigates the impact of modifying the Q-value targets. Specifically, \textbf{pessQ}\citep{wangensemble} represents computing the ensemble Q-value by subtracting an unbiased standard deviation from the mean of the ensemble Q-values, while \textbf{minQ} directly selects the minimum value among the ensemble Q-values. Notably, \textbf{pessQ} does not include gradient diversity optimization, whereas \textbf{minQ} corresponds to the approach used in EDTD7. The relevant hyperparameters are consistent with those listed in \hyperref[tab:environment_hyperparameters]{Table~\ref{tab:environment_hyperparameters}}.

\begin{figure}[H]
    \centering
    \includegraphics[width=1.0\textwidth]{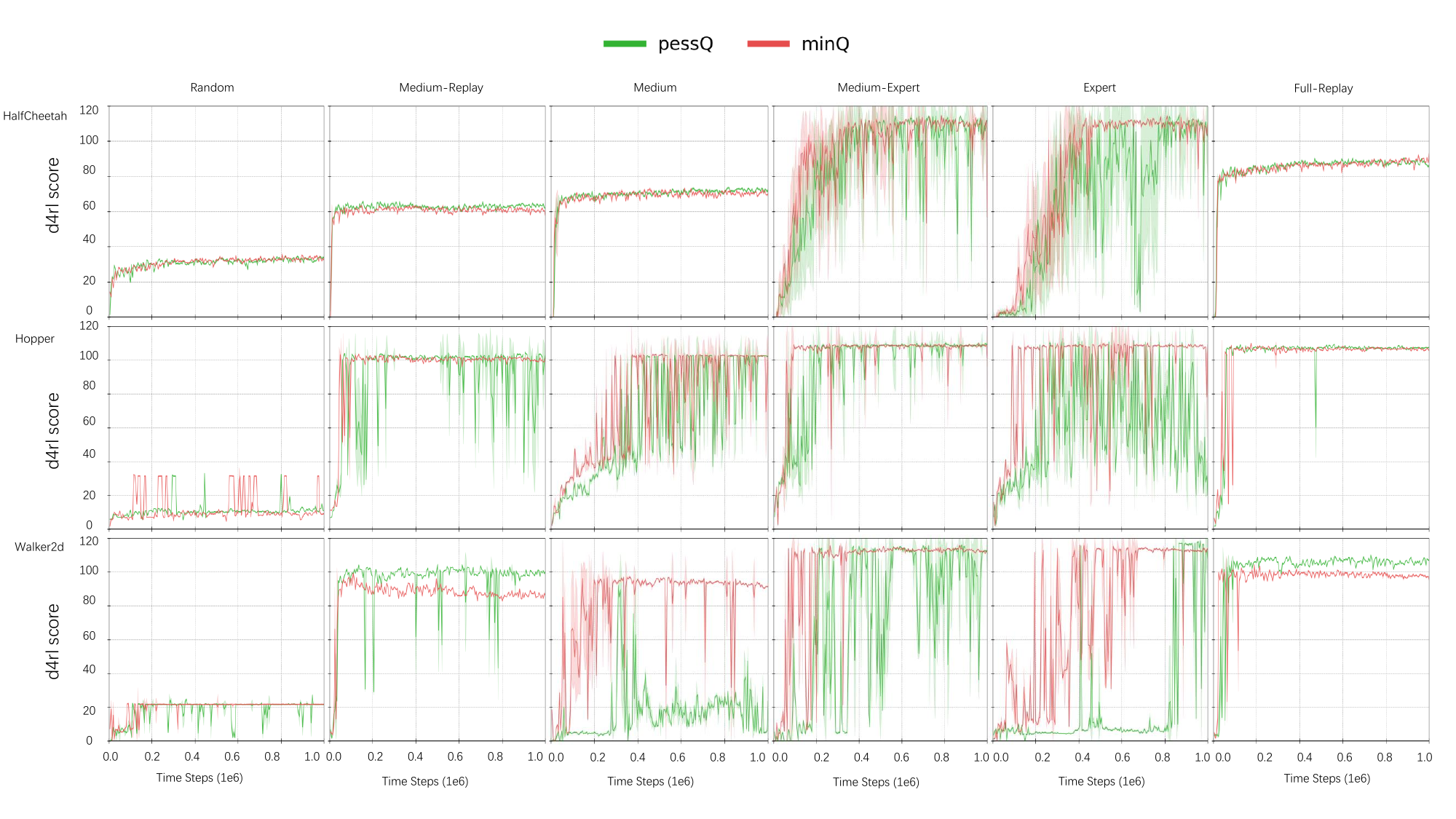}  
    \caption{The learning curves compare the performance of different Q-value targets in EDTD7.  Each curve represents the average performance, and the shaded regions indicate the standard deviation.}
    \label{fig:learning_curves_comparison_q_targets}
\end{figure}

\subsection{Component Ablation}
\label{Ablation}

This subsection conducts a component ablation study, investigating the impact of removing the following components individually: SALE, LAP, and the ensemble network. Notably, removing the ensemble network also eliminates the Q-value gradient diversity penalty term. The remaining hyperparameters are consistent with those listed in \hyperref[tab:environment_hyperparameters]{Table~\ref{tab:environment_hyperparameters}}.

\begin{figure}[H]
    \centering
    \includegraphics[width=1.0\textwidth]{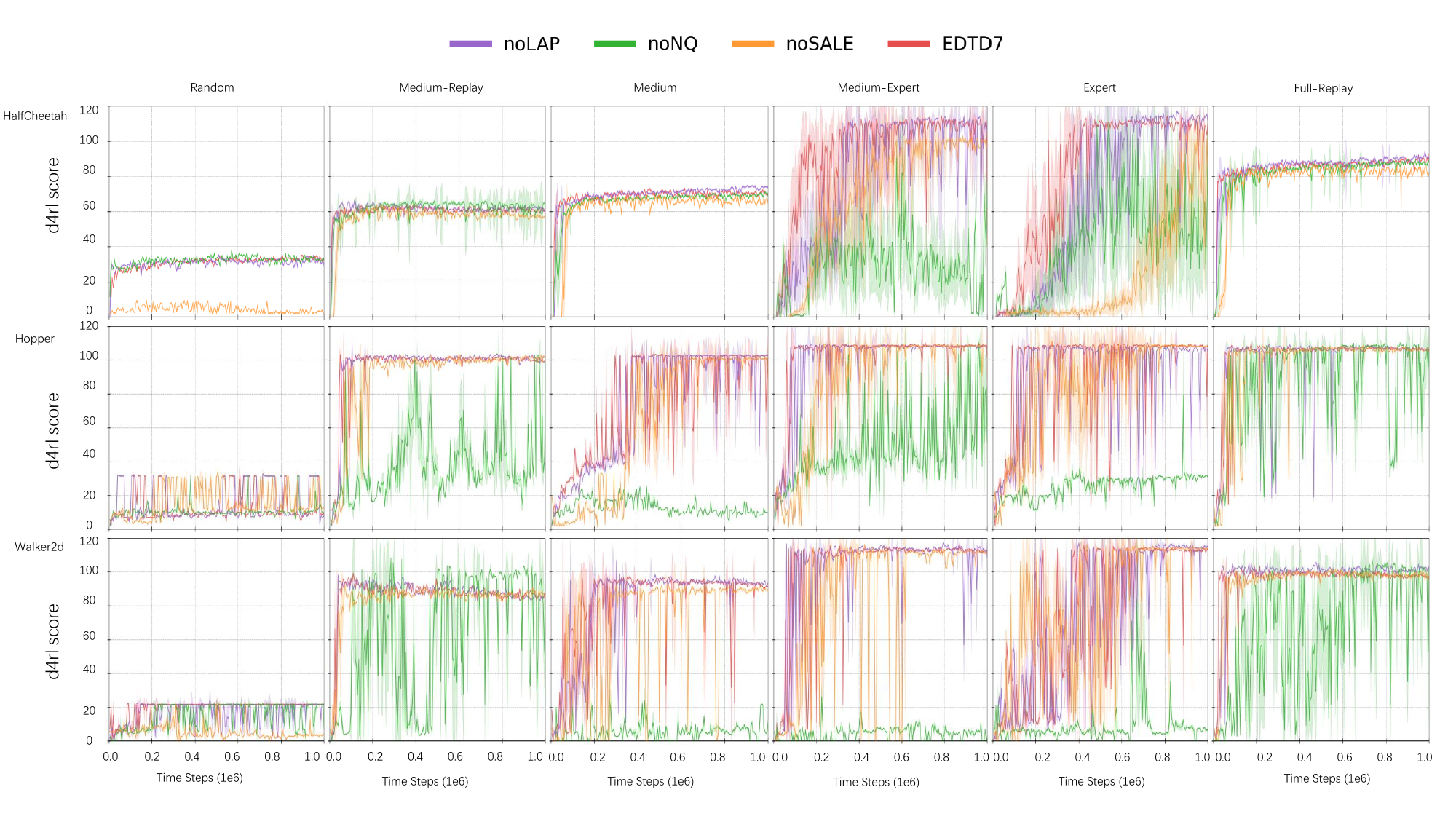}  
    \caption{The learning curves illustrate the ablation results of different components in EDTD7, including SALE, LAP, and the ensemble network. Each curve represents the average performance when a specific component is removed, and the shaded regions indicate the standard deviation.}
    \label{fig:learning_curves_comparison_ablation}
\end{figure}

\end{document}